\documentclass[journal]{IEEEtran}
\IEEEoverridecommandlockouts                              % This command is only
\usepackage{amsmath}
\usepackage{amssymb}
\usepackage{amsfonts}
\usepackage{latexsym}
\usepackage{lipsum}
\usepackage{multicol}
\usepackage{accents}
\usepackage{harpoon}
\usepackage{wrapfig}
\usepackage{tabularx}
\usepackage{euscript}
\usepackage{color}
\usepackage{xcolor}
\usepackage{graphicx}
\usepackage{float}
\usepackage[caption = false]{subfig}	
\usepackage{epsfig}
\usepackage{hyperref}
\usepackage{setspace}
\usepackage{fancyhdr}
\usepackage{textcomp}
\usepackage{cite}
\usepackage{epstopdf}
\usepackage{subfig}
\usepackage{verbatim}
\usepackage[noend]{algpseudocode}
\usepackage{multirow}
\DeclareGraphicsExtensions{.tif,.tiff,.pdf,.eps,.jpg,.png,.PNG,.JPG}
\graphicspath{{Figures/}}

\usepackage{pifont}
\usepackage{latexsym}
\usepackage{psfrag}
\usepackage{stmaryrd}

\newcounter{mynum2}

\usepackage[linesnumbered,ruled,vlined]{algorithm2e}

\providecommand{\DontPrintSemicolon}{\dontprintsemicolon}

\title{\LARGE \bf
A Distributed Control Framework of Multiple Unmanned Aerial Vehicles for Dynamic Wildfire Tracking
}

\author{Huy Xuan Pham, Hung Manh La, David Feil-Seifer, and Matthew Deans% <-this % stops a space
\thanks{Huy Pham is a PhD student, and Dr. Hung La is the director of the Advanced Robotics and Automation
(ARA) Laboratory. Dr. David Feil-Seifer is an Assistant Professor at Department of Computer Science and Engineering, University
of Nevada, Reno, NV 89557, USA. Dr. Matthew Deans is with NASA Ames Research Center, Moffett Field, CA 94035. Corresponding author: Hung La, email: {\tt\small hla@unr.edu}}
}

\begin{document}

\maketitle
\thispagestyle{empty}
\pagestyle{empty}

%%%%%%%%%%%%%%%%%%%%%%%%%%%%%%%%%%%%%%%%%%%%%%%%%%%%%%%%%%%%%%%%%%%%%%%%%%%%%%%%
\begin{abstract}
Wild-land fire fighting is a hazardous job. A key task for firefighters is to observe the ``fire front'' to chart the progress of the fire and areas that will likely spread next. Lack of information of the fire front causes many accidents. Using Unmanned Aerial Vehicles (UAVs) to cover wildfire is promising because it can replace humans in hazardous fire tracking and significantly reduce operation costs. In this paper we propose a distributed control framework designed for a team of UAVs that can closely monitor a wildfire in open space, and precisely track its development. The UAV team, designed for flexible deployment, can effectively avoid in-flight collisions and cooperate well with neighbors. They can maintain a certain height level to the ground for safe flight above fire. Experimental results are conducted to demonstrate the capabilities of the UAV team in covering a spreading wildfire.
\end{abstract}

%%%%%%%%%%%%%%%%%%%%%%%%%%%%%%%%%%%%%%%%%%%%%%%%%%%%%%%%%%%%%%%%%%%%%%%%%%%%%%%%


\section{Introduction}\label{S.intro}

\subsection{Wildfire monitoring and tracking}
\begin{figure}[t]
\centering
\includegraphics[width=0.8\columnwidth]{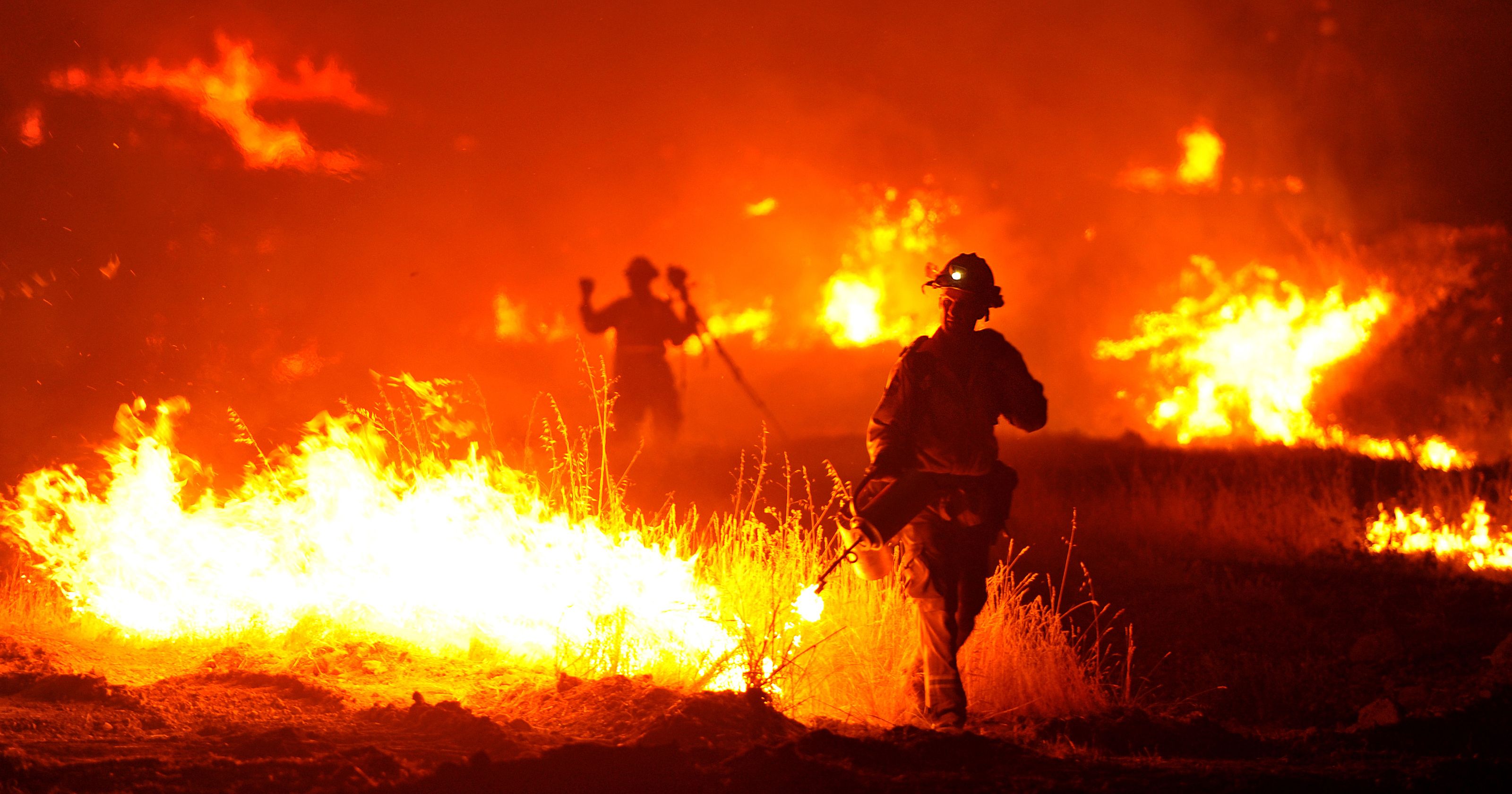}
  \caption{A wildfire outbreaks in California. Firefighting is really dangerous without continuous fire fronts growth information. Courtesy of USA Today.}
  \label{F.Fightfighters}
\vspace{-0pt}
\end{figure}

Wildfire is well-known for their destructive ability to inflict massive damages and disruptions. According  to the U.S. Wildland Fire, an average of 70,000 wildfires annually burn around 7 million acres of land and destroy more than 2,600 structures~\cite{nifc2017}. Wildfire fighting is dangerous and time sensitive; lack of information about the current state and the dynamic evolution of fire contributes to many accidents~\cite{martinez2008computer}. Firefighters may easily lose their life if the fire unexpectedly propagates over them (Figure \ref{F.Fightfighters}). Therefore, there is an urgent need to locate the wildfire correctly~\cite{stipanivcev2010advanced}, and even more important to precisely observe the development of the fire to track its spreading boundaries~\cite{SujitP_2007}. The more information regarding the fire spreading areas collected, the better a scene commander can formulate a plan to evacuate people and property out of danger zones, as well as effectively prevent a fire from spreading to new areas.

Using Unmanned Aircraft Systems (UAS), also called Unmanned Aerial Vehicles (UAV) or drones, to assist wildfire fighting and other natural disaster relief is very promising. They can be used to assist humans for hazardous fire tracking tasks and replace the use of manned helicopters, conserving sizable operation costs in comparison with traditional methods~\cite{MerinoL_2006}~\cite{cruz2016efficient}. However, research that discusses the application of UAVs in assisting fire fighting remains limited~\cite{yuan2015survey}.

UAV technology continues attracting a huge amount of research \cite{Jafari_ISVC2015, Singandhupe_SMC2017}. Researchers developed controllers for UAVs to help them attain stability and effectiveness in completing their tasks \cite{RNC:RNC3687}. Controllers for multirotor UAVs have been thoroughly studied~\cite{luukkonen2011modelling, beard2012small}. In~\cite{woods2017novel, Woods2016_CASE, Woods2015_ISVC}, Wood et al. developed extended potential field controllers for a quadcopter that can track a dynamic target with smooth trajectory, while avoiding obstacles. A Model Predictive Control strategy was proposed in~\cite{bemporad2009hierarchical} for the same objective. UAVs can now host a wide range of sensing capabilities. Accurate UAV-based fire detection has been thoroughly demonstrated in current research. Merino et al.~\cite{MerinoL_2006} proposed a cooperative perception system featuring infrared, visual camera, and fire detectors mounted on different UAV types. The system can precisely detect and estimate fire locations. Yuan et al.~\cite{YuanC_2015} developed a fire detection technique by analyzing fire segmentation in different color spaces. An efficient algorithm was proposed in~\cite{cruz2016efficient} to work on UAV with low-cost cameras, using color index to distinguish fire from smoke, steam and forest environment under fire, even in early stage. Merino et al.~\cite{merino2012unmanned} utilized a team of UAVs to collaborate together to obtain fire front shape and position. In these works, camera plays a crucial role in capturing the raw information for higher level detection algorithms.

\begin{figure*}[!t]
\normalsize
\setcounter{mynum2}{\value{equation}}
\begin{equation}
\label{2.truefarsite}
\begin{aligned}
	X_{t} &= \frac{a^{2}\cos\Theta(x_{s}\sin\Theta + y_{s}\cos\Theta) - b^{2}\sin\Theta(x_{s}\cos\Theta - y_{s}\sin\Theta)}{\sqrt{b^{2}(x_{s}\cos\Theta + y_{s}\sin\Theta) - a^{2}(x_{s}\sin\Theta - y_{s}\cos\Theta}}  
		+ c\sin\Theta) \\
	Y_{t} &= \frac{-a^{2}\sin\Theta(x_{s}\sin\Theta + y_{s}\cos\Theta) - b^{2}\cos\Theta(x_{s}\cos\Theta - y_{s}\sin\Theta)}{\sqrt{b^{2}(x_{s}\cos\Theta + y_{s}\sin\Theta) - a^{2}(x_{s}\sin\Theta - y_{s}\cos\Theta}}  
		+ c\cos\Theta),\\
\end{aligned}
\end{equation}
\setcounter{equation}{1}
\hrulefill
\vspace{-15pt}
\end{figure*}

\subsection{Multiple robots in sensing coverage}

Using multiple UAVs as a sensor network \cite{Cui_SMCA_2016}, especially in hazardous environment or disaster, is well discussed. In~\cite{La_CASE2013, La_SMCA_2015}, La et al. demonstrated how multiple UAVs can reach consensus to build a scalar field map of oil spills or fire. Maza et al.~\cite{MazaI_2011} provided a distributed decision framework for multi-UAV applications in disaster management. Specific applications in wildfire monitoring involving multiple robots systems have been reported. In~\cite{CasbeerD_2005}, multiple UAVs are commanded to track a spreading fire using checkpoints calculated based on visual images of the fire perimeter. Artificial potential field algorithms have been employed to control a team of UAVs in two separated tasks: track the boundary of a wildfire and suppress it~\cite{kumar2011cooperative}. A centralized optimal task allocation problem has been formulated in~\cite{phan2008cooperative} to generate a set of waypoints for UAVs for shortest path planning.

However, to the best of the authors' knowledge, most of the above mentioned work does not cover the behaviors of their system when the fire is spreading. Works in~\cite{CasbeerD_2005} and~\cite{phan2008cooperative} centralized the decision making, thus potentially overloaded in computation and communication when the fire in large scale demands more UAVs. The team of UAVs in~\cite{kumar2011cooperative} can continuously track the boundary of the spreading fire but largely depends on the accuracy of the modeled shape function of the fire in control design. 

Knowledge on optimal sensing coverage using multiple, decentralized robots \cite{Dang_MFI2016, TNguyen_Allerton2016} could be applied to yield better results in wildfire monitoring. This itself is a large, active research branch. Cortes et al. in~\cite{cortes2004coverage} categorized the optimal sensor coverage problem as a locational optimization problem. Schwager et al. in~\cite{schwager2009decentralized} presented a control law for a team of networked robots using Voronoi partitions for a generalized coverage problem. Subsequent works such as~\cite{breitenmoser2010voronoi} and~\cite{adibi2013adaptive} expanded to work with non-convex environment with obstacles. In~\cite{pimenta2009simultaneous}, the coverage problem was expanded to also detect and track moving targets within a fixed environment. Most of the aforementioned works only considered the coverage of a fixed, static environment, while the problem in wildfire coverage requires a framework that can work with a changing environment.

In this paper, we characterize the optimal sensing coverage problem to work with a changing environment. We propose a decentralized control algorithm for a team of UAVs that can autonomously and actively track the fire spreading boundaries in a distributed manner, without dependency on the wildfire modeling. The UAVs can effectively share the vision of the field, while maintaining safe distance in order to avoid in-flight collision. Moreover, during tracking, the proposed algorithm can allow the UAVs to increase image resolution captured on the border of the wildfire. This idea is greatly inspired by the work of Schwager et al. in~\cite{schwager2011eyes}, where a decentralized control strategy was developed for a team of robotic cameras to minimize the information loss over an environment. For safety reason, our proposed control algorithm also allows each UAV to maintain a certain height level to the ground to avoid getting caught by the fire. Note that, the initial results of this study were published in the conference proceedings~\cite{8206579}.

The rest of the paper is organized as follows: Section 2 discusses how wildfire spreading is modeled as an objective for this paper. In Section 3, the wildfire tracking problem is formulated with clear objectives. In Section 4, we propose a control design capable of solving the problem. A simulation scenario on MATLAB are provided in Section 5. Finally, we draw a conclusion, and suggest directions for future work.

\begin{figure*}[t]
 \centering
	\subfloat[t = 0]{\includegraphics[width=0.25\textwidth]{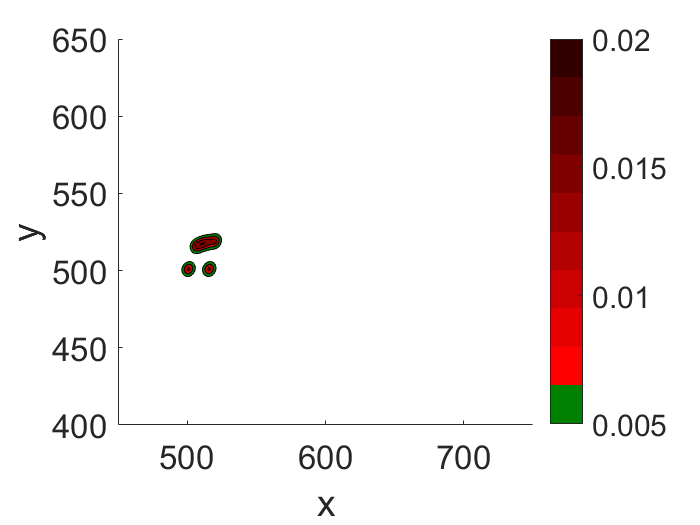}}
	\subfloat[t = 1000]{\includegraphics[width=0.25\textwidth]{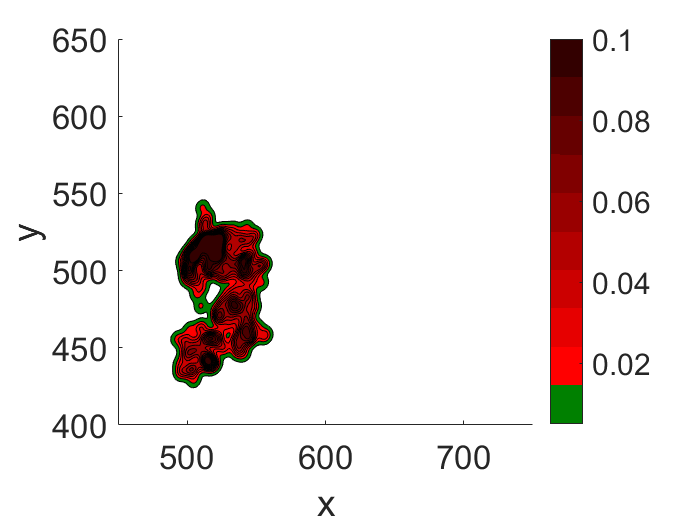}}
	\subfloat[t= 3000]{\includegraphics[width=0.25\textwidth]{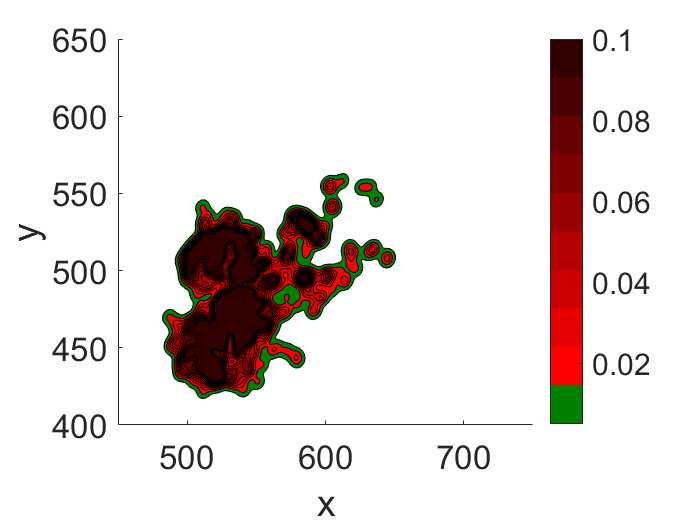}}
	\subfloat[t = 6000]{\includegraphics[width=0.25\textwidth]{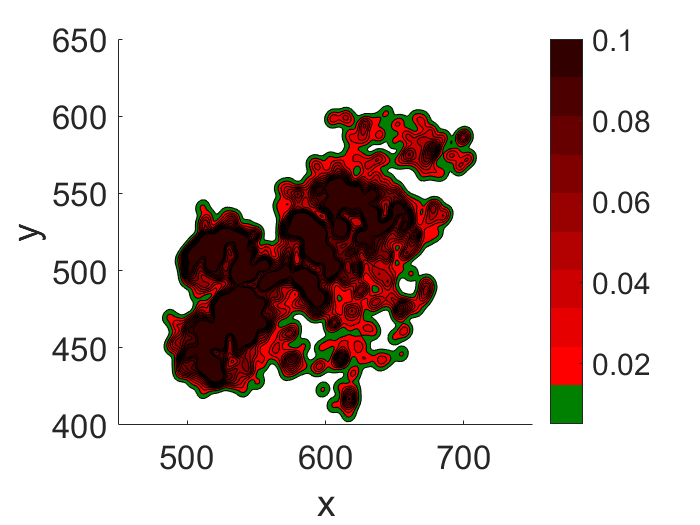}}
  \caption{Simulation result shows a wildfire spreading at different time steps. The wildfire starts with a few heat sources around $(500, 500)$, grows bigger and spreads around the field. The color bar indicates the intensity level of the fire. The darker the color, the higher the intensity.}
  \label{F.wildfire}
\vspace{-0pt}
\end{figure*}

\section{Wildfire Modeling}\label{S.wildfire}

Wildfire simulation has attracted significant research efforts over the past decades, due to the potential in predicting wildfire spreading. The core model of existing fire simulation systems is the fire spreading propagation~\cite{glasa2011note}. Rothermel in 1972~\cite{rothermel1972mathematical} developed basic fire spread equations to mathematically and empirically calculate rate of speed and intensity. Richards~\cite{richards1990elliptical} introduced a technique to estimate fire fronts growth using an elliptical model. These previous research were later developed further by Finney~\cite{finney2004farsite} and became a well-known fire growth model called Fire Area Simulator (FARSITE). Among existing systems, FARSITE is the most reliable model~\cite{williams2014modeling}, and widely used by federal land management agencies such as U.S. Department of Agriculture (USDA) Forest Service. However, in order to implement the model precisely, we need significant information regarding geography, topography, conditions of terrain, fuels, and weather. To focus on the scope of multi-UAV control rather than pursuing an accurate fire growth model, in this paper we modify the fire spreading propagation in FARSITE model to describe the fire front growth in a simplified model. We make the following assumptions:

\begin{itemize}
\item the model will be implemented for a discrete grid-based environment;
\item the steady-state rate of spreading is already calculated for each grid;
\item only the fire front points spread.
\end{itemize}

Originally, the equation for calculating the differentials of spreading fire front proposed in~\cite{richards1990elliptical} and~\cite{finney2004farsite} as Equation (\ref{2.truefarsite}), where $X_{t}$ and $Y_{t}$ are the differentials, $\Theta$ is the azimuth angle of the wind direction and $y$-axis ($0\leq \Theta \leq 2\pi$). $\Theta$ increases following clock-wise direction. $a$ and $b$ are the length of semi-minor and semi-major axes of the elliptical fire shape growing from one fire front point, respectively. $c$ is the distance from the fire source (ignition point) to the center of the ellipse. $x_{s}$ and $y_{s}$ are the orientation of the fire vertex. We simplify the Equation (\ref{2.truefarsite}) to only retain the center of the new developed fire front as follows:
\begin{equation}\label{2.firespread}
\begin{aligned}
	X_{t} &= c\sin\Theta \\
	Y_{t} &= c\cos\Theta. \\ 
\end{aligned}
\end{equation}
We use equation from Finney~\cite{finney2004farsite} to calculate $c$ according to the set of equations (\ref{2.firespread}) as follows:
\begin{equation}\label{E.rel_distance}
\begin{aligned}
	\ c &= \frac{R - \frac{R}{HB}}{2}\\
	\ HB &= \frac{LB + (LB^{2} - 1)^{0.5}}{LB - (LB^{2} - 1)^{0.5} }\\
	\ LB &= 0.936 e^{0.2566U} + 0.461 e^{- 0.1548U}  - 0.397,\\	
\end{aligned}
\end{equation}
where $R$ is the steady-state rate of fire spreading. $U$ is the scalar value of mid-flame wind speed, which is the wind speed at the ground. It can be calculated from actual wind speed value after taking account of the wind resistance by the forest. The new fire front location after time step $\delta t$ is calculated as:
\begin{equation}\label{2_location}
\begin{aligned}
	x_{f}(t+\Delta t) &= x_{f}(t) + \Delta tX_{t}(t)\\
	y_{f}(t+\Delta t) &= y_{f}(t) + \Delta tY_{t}(t).\\
\end{aligned}
\end{equation}
Additionally, in order to simulate the intensity caused by fire around each fire front source, we also assume that each fire front source would radiate energy to the surrounding environment resembling a multivariate normal distribution probability density function of its coordinates $x$ and $y$. Assuming linearity, the intensity of each point in the field is a linear summation of intensity functions caused by multiple fire front sources. Therefore, we have the following equation describing the intensity of each point in the wildfire caused by a number of $k$ sources:
\begin{equation}\label{2.multivariate}
\begin{aligned}
	\ I(x, y) &= \sum^{k}_{i = 1}\frac{1}{2\pi\sigma_{x_{i}}\sigma_{y_{i}}}e^{-\frac{1}{2}[\frac{(x-x_{f})^2}{\sigma_{x_{i}}^2}+\frac{(y-y_{f})^2}{\sigma_{y_{i}}^2}]}, \\
\end{aligned}
\end{equation}
where $I(x, y)$ is the intensity of the fire at a certain point $q(x, y)$, $(x_{f}, y_{f})$ is the location of the heat source $i$, and $(\sigma_{x_{i}}, \sigma_{y_{i}})$ are deviations. The point closer to the heat source has a higher level of intensity of the fire. Figure \ref{F.wildfire} represents the simulated wildfire spreading from original source (a) until $t = 6000$ time steps (d). The simulation assumes the wind flows north-east with direction is normally distributed ($\mu_{\Theta} = \frac{\pi}{8}, \sigma_{\Theta} = 1$), midflame adjusted wind speed is also normally distributed ($\mu_{U} = 5, \sigma_{u} = 2$). The green area depicts the boundary with forest field, while red area represents the fire. The brighter red color area illustrates the outer of the fire and regions near the boundary where the intensity is lower. The darker red colors show the area in fire with high intensity.

It should be noted that in this paper, the accuracy of the model should not affect the performance of our distributed control algorithm, as explained in section IV, subsection A. In case a different model of wildfire spreading is used, for instance, by changing equation set (\ref{2.firespread}) and (\ref{E.rel_distance}), only the shape of the wildfire changes, but the controller should still work. 

\section{Problem formulation}\label{S.formulation}

In this section, we translate our motivation into a formal problem formulation. Our objective is to control a team of multiple UAVs for collaboratively covering a wildfire and tracking the fire front propagation. By covering, we mean to let the UAVs take multiple sub-pictures of the affected area so that most of the entire field is captured. We assume that the fire happens in a known section of a forest, where the priori information regarding the location of any specific point are made available. Suppose that when a wildfire happens, its estimated location is notified to the UAVs. A command is then sent to the UAV team allowing them to start. The team needs to satisfy the following objectives:

\begin{itemize}
\item \textit{Deployment objective}: The UAVs can take flight from the deployment depots to the initially estimated wildfire location. 
\item \textit{Coverage and tracking objective}: Upon reaching the reported fire location, the team will spread out to cover the entire wildfire from a certain altitude. The UAVs then follow and track the development of the fire fronts. When following the expanding fire fronts of the wildfire, some of the UAV team may lower their altitude to increase the image resolution of the fire boundary, while the whole team tries to maintain a complete view of the wildfire.
\item \textit{Collision avoidance and safety objective}: Because the number of UAVs can be large (i.e. for sufficient coverage a large wildfire), it is important to ensure that the participating UAVs are able to avoid in-flight collisions with other UAVs. Moreover, a safety distance between the UAVs and the ground should be established to prevent the UAVs from catching the fire.
\end{itemize}

\begin{figure}[h]
 \centering
  \includegraphics[width=0.8\columnwidth]{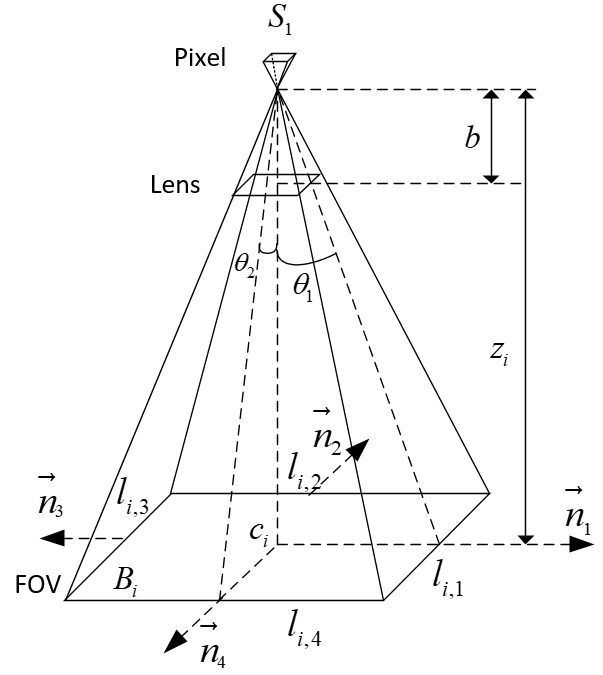}
  \caption{Rectangular FOV of a UAV, with half-angles $\theta_{1}$, $\theta_{2}$, composing from 4 lines $l_{i,1}, l_{i,2}, l_{i,3}, l_{i,4}$ and their respective normal vector $n_{1}, n_{2}, n_{3}, n_{4}$. Each UAV will capture the area under its field of view using its camera, and record the information into a number of pixels.}
  \label{F.fieldofview}
\vspace{-0pt}
\end{figure}
\noindent

\begin{figure}[h]
\centering
  \includegraphics[width=0.8\columnwidth]{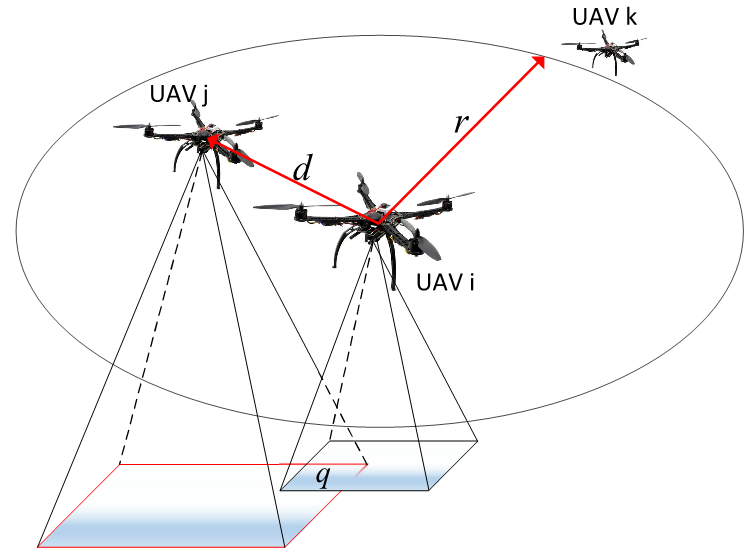}
  \caption{UAV $i$ only communicates with a nearby UAV inside its communication range $r$ (UAV $j$) (their \textit{physical neighbor}). Each UAV would try to maintain a designed safe distance $d$ to other UAVs in the team. If two physical neighboring UAVs cover one common point $q$, they are also \textit{sensing neighbors}.}
  \label{F.potentialfieldneighbors}
\vspace{-0pt}
\end{figure}

Assume that each UAV equipped with localization devices (such as GPS and IMU), and identical downward-facing cameras capable of detecting fire. Each camera has a rectangular \textit{field of view} (FOV). When covering, the camera and its FOV form a pyramid with half-angles $\theta^{T} =[\theta_{1}, \theta_{2}]^{T}$ (see Figure \ref{F.fieldofview}). Each UAV will capture the area under its FOV using its camera, and record the information into an array of pixels. We also assume that a UAV can communicate and exchange information with other UAVs if it remains inside a communication sphere with radius $r$ (see Figure \ref{F.potentialfieldneighbors}). 

%Figure 4: Radius r
%\noindent

We define the following variables that will be used throughout this paper.
Let $N$ denote the set of the UAVs. Let $p_{i} = [c_{i}^{T}, z_{i}]^{T}$denote the pose of a UAV $i \in N$. In which, $c_{i}^{T} = [x_{i}, y_{i}]^{T}$ indicates the lateral coordination, and $z_{i}$ indicates the altitude. Let $B_{i}$ denote the set of points that lie inside the field of view of UAV $i$. Let $l_{k,i}, k = 1:4$ denotes each edge of the rectangular FOV of UAV $i$. Let $n_{k}, k = 1:4$ denotes the outward-facing normal vectors of each edge, where $n_{1} = [1, 0]^{T}$,  $n_{2} = [0, 1]^{T}$,  $n_{3} = [-1, 0]^{T}$,  $n_{4} = [0, -1]^{T}$. We then define the objective function for each task of the UAV team.

\subsection{Deployment objective}
The UAVs can be deployed from depots distributed around the forest, or from a forest firefighting department center. Upon receiving the report of a wildfire, the UAVs are commanded to start and move to the point where the location of the fire was initially estimated. We call this point a rendezvous point $p_{r} = [p_{x}, p_{y}, p_{z}]^{T}$. The UAVs would keep moving toward this point until they can detect the wildfire inside their FOV.

\subsection{Collision avoidance and safety objective}
The team of UAVs must be able to avoid in-flight collision. In order to do that, a UAV needs to identify its neighbors first. UAV $i$ only communicates with a nearby UAV $j$ that remains inside its communication range (Figure \ref{F.potentialfieldneighbors}), and satisfies the following equation:
\begin{equation}\label{3_flockneighbors}
\begin{aligned}
	\ ||p_{j} - p_{i}|| \leq r, \\
\end{aligned}
\end{equation}
where $||.||$ is Euclidean distance, and $r$ is the communication range radius. If Equation (\ref{3_flockneighbors}) is satisfied, the two UAVs become \textit{physical neighbors}. For UAV $i$ to avoid collision with other neighbor UAV $j$, they must keep their distance not less than a designed distance $d$:
\begin{equation}\label{3_avoidcollision}
\begin{aligned}
	\ ||p_{j} - p_{i}|| \geq d. \\
\end{aligned}
\end{equation}
As we proposed earlier, during the implementation of the tracking and coverage task, the UAVs can lower their altitude to increase the resolution of the border of the wildfire. Since there is no obvious guarantee about the minimum altitude of the UAVs, they can keep lowering their altitude, and may catch fire during their mission. Therefore, it is imperative that the UAVs must maintain a safe distance to the ground. Suppose the safe altitude is $z_{min}$, and infer the position of the image of the UAV $i$ as $p_{i^{\prime}} = [c_{i}^{T}, 0]$, we have the safe altitude condition:
\begin{equation}\label{3_avoidground}
\begin{aligned}
	\ ||p_{i} - p_{i^{\prime}}|| \geq z_{min}. \\
\end{aligned}
\end{equation}

\subsection{Coverage and tracking objective}
Let $Q(t)$ denote the wildfire varying over time $t$ on a plane. The general optimal coverage problem is normally represented by a coverage objective function with the following form:
\begin{equation}\label{3_generalobjective}
\begin{aligned}
	\min H(p_{1},...,p_{n}) &= \int_{Q(t)}f(q,p_{1},p_{2},...,p_{n})\phi(q,t)dq,
\end{aligned}
\end{equation}
\noindent
where $f(q,p_{1},p_{2},...,p_{n})$ represents some cost to cover a certain point $q$ of the environment. The function $\phi(q, t)$, which is known as distribution density function, level of interestingness, or strategic importance, indicates the specific weight of the point $q$ in that environment at time $t$. In this paper, the cost we are interested in is the quality of images when covering a spreading fire with a limited number of cameras. This notion was first described in~\cite{schwager2011eyes}. Since each camera has limited number of pixels to capture an image, it will provide one snapshot of the wildfire with lower resolution when covering it in a bigger FOV, and vice versa. By minimizing the information captured by the pixels of all the cameras, in other word, the area of the FOVs containing the fire, we could provide with optimal-resolution images of the fire.

To quantify the cost, we first consider the image captured by one camera. Digital camera normally uses photosensitive electronics which can host a large number of pixels. The quality of recording an image by a single pixel can represent the quality of the image captured by that camera. From the relationship between object and image distance through a converging lens in classic optics, we can easily calculate the FOV area that a UAV covers (see figure \ref{F.fieldofview}) as follows:
\begin{equation}\label{3_areaperpixel}
\begin{aligned}
	\ f(q, p_{i}) = \frac{S_{1}}{b^{2}}(b-z_{i})^{2}, \forall q \in B_{i}, \\
\end{aligned}
\end{equation}
\noindent
where $q^{T} = [q_{x}, q_{y}]^{T}$ is the coordination of a given point that belongs to $Q(t)$, $S_{1}$ is the area of one pixel of a camera, and $b$ denotes the focal length. Note that, for a point $q$ to lie on or inside the FOV of a UAV $i$, it must satisfy the following condition:
\begin{equation}\label{3_coveragecondition}
\begin{aligned}
	\frac{||q - c_{i}||}{z_{i}} \leq \tan \theta. \\
\end{aligned}
\end{equation}
From Equation (\ref{3_areaperpixel}), it is obvious that the higher the altitude of the camera ($z_{i}$) is, the higher the cost the camera incures, or the lower its image resolution is. 

For multiple cameras covering a point $q$, Schwager et al.~\cite{schwager2011eyes} formulated a cost to represent the coverage of a point $q$ in a static field $Q$ over total number of pixels from a multiple of $n$ cameras as follows:
\begin{equation}\label{3_Schwagersobjectiveoriginal}
\begin{aligned}
	f_{ N_{q}}(q, p_{1},...,p_{n}) &= (\sum_{i\in N_{q}}f(p_{i},q)^{-1})^{-1},
\end{aligned}
\end{equation}
\noindent
where $f(p_{i},q)$ calculated as in equation (\ref{3_areaperpixel}), $N_{q}$ is the set of UAVs that include the point $q$ in their FOVs. However, in case the point $q$ is not covered by any UAV, $f(p_{i},q) = \infty$, the denominator in (\ref{3_Schwagersobjectiveoriginal}) can be come zero. To avoid zero division, we need to introduce a constant $m$: 
\begin{equation}\label{3_Schwagersobjective}
\begin{aligned}
	f_{ N_{q}}(q, p_{1},...,p_{n}) &= (\sum_{i\in N_{q}}f(p_{i},q)^{-1}+m)^{-1}.
\end{aligned}
\end{equation}
The value of $m$ should be very small, so that in such case, the cost in (\ref{3_Schwagersobjective}) become very large, thus discouraging this case to happen. We further adapt the objective function (\ref{3_generalobjective}) so that the UAVs will try to cover the field in the way that considers the region around the border of the fire more important. First, we consider that each fire front radiates a heat aura, as described in Equation (\ref{2.multivariate}), Section \ref{S.wildfire}. The border region of each fire front has the least heat energy, while the center of the fire front has the most intense level. We assume that the UAVs equipped with infrared camera allowing them to sense different color spectra with respect to the levels of fire heat intensity. Furthermore, the UAVs are assumed to have installed an on-board fire detection program to quantify the differences in color into varying levels of fire heat intensity~\cite{cruz2016efficient}. Let $I(q)$ denote the varying levels of fire heat intensity at point $q$, and suppose that the cameras have the same detection range $[I_{min}, I_{max}]$. The desired objective function that weights the fire border region higher than at the center of the fire allows us to characterize the importance function as follows:
\begin{equation}\label{3_phi}
\begin{aligned}
	\phi(q) &= \kappa (I_{max} - I(q)) &= \kappa \Delta I(q).\\
\end{aligned}
\end{equation}
One may notice that the intensity $I(q)$ actually changes over time. This makes $\phi(q)$ depends on the time, and would complicate Equation (\ref{3_generalobjective})~\cite{pimenta2009simultaneous}. In this paper, we assume that the speed of the fire spreading is much less than the speed of the UAVs, therefore at a certain period of time, the intensity at a point can be considered constant. Also, note that some regions at the center of the wildfire may have $I = I_{max}$ now become not important. This makes sense because these regions likely burn out quickly, and they are not the goals for the UAV to track. We have the following objective function for wildfire coverage and tracking objective:
\begin{equation}\label{3_coverobjective}
\begin{aligned}
	\min H &= \int_{Q(t)}(\sum_{i\in N_{q}}f(p_{i},q)^{-1}+ m)^{-1}\kappa \Delta I(q)dq.\\
\end{aligned}
\end{equation}
Note that when two UAVs have one or more points in common, they will become \textit{sensing neighbors}. For a UAV to identify the set $N_{q}$ of a point $q$ inside its FOV, that UAV must know the pose of other UAVs as indicated by the condition (\ref{3_coveragecondition}). Therefore, in order to become sensing neighbors, the UAVs must first become physical neighbors, defined by (\ref{3_flockneighbors}). One should notice this condition to select the range radius $r$ large enough to guarantee communication among the UAVs that have overlapping field of views. But we must also limit $r$ so that communication overload does not occur as a result of having too many neighbors.

\section{Controller Design}\label{S.control}

\begin{figure}[t]
 \centering
  \includegraphics[width=0.8\columnwidth]{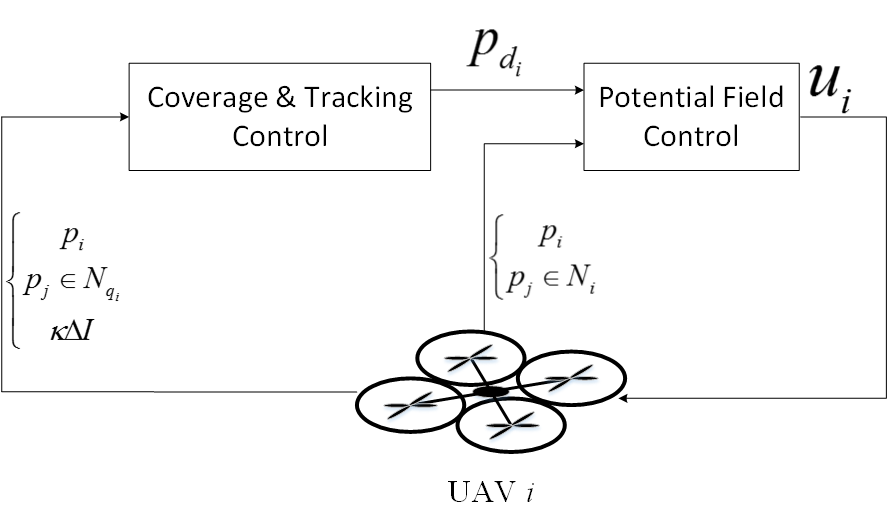}
  \caption{Controller architecture of UAV $i$, consisting of two components:  the Coverage and Tracking component and the Potential Field component. The Coverage and Tracking component generates the desired position, $p_{d_{i}}$, for the UAV for wildfire coverage and tracking. The Potential Field component controls the UAV to move to the desired positions, which were  generated by the Coverage \& Tracking component, and to avoid collision with other UAVs and the ground.}
  \label{F.Controller}
\vspace{-0pt}
\end{figure}

Figure \ref{F.Controller} shows our controller architecture for each UAV. Our controller consists of two components: the coverage and tracking component and the potential field component. The coverage and tracking component calculates the position of the UAV for wildfire coverage and tracking. The potential field component controls the UAV to move to desired positions, and to avoid collision with other UAVs, as well as maintain the safety distance to the ground, by using potential field method.  Upon reaching the wildfire region, the coverage and tracking control component will update the desired position of the UAV to the potential field control component. Assume the UAVs are quadcopters, then the dynamics of each UAV is:
\begin{equation}\label{4.dynamics}
\begin{aligned}
	\ u_{i} &= \dot{p_{i}}, \\
\end{aligned}
\end{equation}
\noindent
we can then develop the control equation for each component in the upcoming subsections.

\subsection{Coverage \& tracking control}
Based on the artificial potential field approach~\cite{schwager2011eyes,ge2000new, La_RAS_2012}, each UAV is distributedly controlled by a negative gradient (gradient descent) of the objective function $H$ in equation (\ref{3_coverobjective}) with respect to its pose $p_{i} = [c_{i}, z_{i}]^{T}$ as follows:
\begin{equation}\label{4.gradientcontrol}
\begin{aligned}
	\ u_{i}^{ct} &= -k_{s}\frac{\partial H}{\partial p_{i}}, \\
\end{aligned}
\end{equation}
where $k_{s}$ is the proportional gain parameter. Taking the derivative with notation that $Q(t) = (Q(t) \cap B_{i}) \cup (\partial{(Q(t) \cap B_{i})}) \cup (Q(t) \setminus B_{i}) \cup (\partial{(Q(t) \setminus B_{i})})$ as in~\cite{schwager2011eyes}, where $\partial{.}$ denotes the boundaries of a set, we have:
\begin{equation}\label{4.gradientcontrol_step1}
\begin{aligned}
	\ \frac{\partial H}{\partial p_{i}} &= \frac{\partial}{\partial p_{i}} \int \limits_{Q(t) \cap B_{i}} f_{N_{q}}\Delta I dq + \frac{\partial}{\partial p_{i}}\int \limits_{\partial{(Q(t) \cap B_{i})}} f_{N_{q}}\Delta I dq \\
	&+ \frac{\partial}{\partial p_{i}}\int \limits_{\partial{(Q(t) \setminus B_{i})}} f_{N_{q \setminus i}}\Delta I dq + \frac{\partial}{\partial p_{i}}\int \limits_{Q(t) \setminus B_{i}} f_{N_{q \setminus i}}\Delta I dq. \\
\end{aligned}
\end{equation}
In the last component, $Q(t) \setminus B_{i}$ does not depend on $p_{i}$ so it is equal to zero. Then the lateral position and altitude of each UAV is controlled by taking the partial derivatives of the objective function $H$ as follows:
\begin{equation}\label{4.gradientcomponent}
\begin{aligned}
	\frac{\partial H}{\partial c_{i}} &= \sum_{k=1}^{4}\int \limits_{Q(t) \cap l_{k,i}}(f_{N_{q}} - f_{N_{q}\setminus i})n_{k}\kappa\Delta Idq, \\
	\frac{\partial H}{\partial z_{i}} &= \sum_{k=1}^{4}\int \limits_{Q(t) \cap l_{k,i}}(f_{N_{q}} - f_{N_{q}\setminus i})\tan \theta^{T}n_{k}\kappa\Delta Idq,	\\
	&- \int \limits_{Q(t) \cap B_{i}} \frac{2f_{N_{q}}^{2}}{ \frac{S_{1}}{b^{2}}(b-z_{i})^{3}}\kappa\Delta Idq,   
\end{aligned}
\end{equation}
where $f_{N_{q}}$ and $ f_{N_{q\setminus i}}$ are calculated as in equation (\ref{3_Schwagersobjective}), $N_{q}\setminus{i}$ denotes the coverage neighbor set excludes the UAV $i$. In (\ref{4.gradientcomponent}), the component $\frac{\partial H}{\partial c_{i}}$ allows the UAV to move along $x$-axis and $y$-axis of the wildfire area which has $\Delta I$ is larger, while reducing the coverage intersections with other UAVs. The component $\frac{\partial H}{\partial z_{i}}$ allows the UAV to change its altitude along the $z$-axis to trade off between cover larger FOV (the first component) over the wildfire and to have a better resolution of the fire fronts propagation (the second component). This set of equations is similar to the one proposed in~\cite{schwager2011eyes}, except that we extend them to work with an environment $Q(t)$, which now changes over the time, and the weight function $\phi(q)$ is characterized specifically to solve the dynamic wildfire tracking problem.

In order to compute these control inputs in (\ref{4.gradientcomponent}), one needs to determine $Q(t) \cap B_{i}$ and $Q(t) \cap l_{k,i}$. This can be done by discretize $B_{i}$ (i.e. area inside the FOV) and $l_{k,i}$ (i.e. the edges of the FOV) of a UAV $i$ into discrete points, and check if those points also belong to $Q$ at time $t$, or in other words, check the level of intensity of each point by using the fire detection system of the UAV. Obviously, we need to assume the intensity model of the environment in (\ref{2.multivariate}) to hold true, hence our approach is still model-based. However, we would not need explicit information such as the accurate shape of the fire, as in~\cite{kumar2011cooperative}, to implement the controller. This is an advantage, since it is more difficult to get an accurate shape model of the fire, comparing to the reasonable assumption of fire intensity model.

From (\ref{4.gradientcomponent}), the desired virtual position $p_{d_{i}}$ will be updated to the potential field control component (see Figure \ref{F.Controller}):
\begin{equation}\label{4.virtualposition}
\begin{aligned}
	\ p_{d_{i}}(t+\Delta t) &= p_{d_{i}}(t) - u_{i}^{ct}\Delta t, u_{i}^{ct} = (k_{c}\frac{\partial H}{\partial c_{i}}, k_{z}\frac{\partial H}{\partial z_{i}}).\\
\end{aligned}
\end{equation}

\subsection{Potential field control}

The objective of this component is to control a UAV from the current position to a new position updated from the coverage and tracking control. Similarly, our approach is to create an artificial potential field to control each UAV to move to a desired position, and to avoid in-flight collision with other UAVs. We first create an attractive force to pull the UAVs to the initial rendezvous point $p_{r}$ by using a quadratic function of distance as the potential field, and take the gradient of it to yield the attractive force:
\begin{equation}\label{4.initialattractive}
\begin{aligned}
	\ U_{r}^{att} &= \frac{1}{2}k_{r}||p_{r} - p_{i}||^{2}\\
	\ u_{i}^{r} &= - \nabla U_{r}^{att} = -k_{r}(p_{i} - p_{r}). \\
\end{aligned}
\end{equation}
Similarly, the UAV moves to desired virtual position, $p_{d_{i}}$, passed from equation (\ref{4.virtualposition}) in coverage \& tracking component, by using this attractive force:
\begin{equation}\label{4.virtualattractive}
\begin{aligned}
	\ U_{d}^{att} &= \frac{1}{2}k_{d}||p_{d_{i}} - p_{i}||^{2}\\
	\ u_{i}^{d} &= - \nabla U_{d}^{att} = -k_{d}(p_{i} - p_{d_{i}}). \\
\end{aligned}
\end{equation}
In order to avoid collision with its neighboring UAVs, we create repulsive forces from neighbors to push a UAV away if their distances become less than a designed safe distance $d$. Define the potential field for each neighbor UAV $j$ as:
\begin{equation}\label{4.repulsivePF}
\begin{aligned}
	\ U_{j}^{rep} &= 
	\begin{cases}
	\ \frac{1}{2}\nu(\frac{1}{||p_{j} - p_{i}||} - \frac{1}{d})^{2}, & if  \  ||p_{j} - p_{i}|| < d \\
    	 \ 0, & otherwise, \\
    	\end{cases}
\end{aligned}
\end{equation}
\noindent
where $\nu$ is a constant. The repulsive force can be attained by taking the gradient of the sum of the potential fields created by all neighboring UAVs as follows:
\begin{equation}\label{4.repulsive}
\begin{aligned}
	\ u_{i}^{rep1} &= - \sum_{j \in N_{i}}a_{ij}\nabla U_{j}^{rep} \\
	&=-\sum_{j \in N_{i}}\nu a_{ij}\Big(\frac{1}{||p_{j} - p_{i}||} - \frac{1}{d}\Big)\frac{1}{||p_{j}-p_{i}||^{3}}(p_{i} - p_{j})\\
	a_{ij} &=
	\begin{cases}
	\  1, &  if  \  ||p_{j} - p_{i}|| < d \\
    	 \ 0, &  otherwise. \\
    	\end{cases}
\end{aligned}
\end{equation}
Similarly, for maintaining a safe distance to the ground, we have:
\begin{equation}\label{4.altitude_repulsive}
\begin{aligned}
	\ u_{i}^{rep2} &= - a_{ii^{\prime}}\nabla U_{i^{\prime}}^{rep} \\
	&=-\nu^{\prime} a_{ii^{\prime}}\Big(\frac{1}{||p_{i^{\prime}} - p_{i}||} - \frac{1}{z_{min}}\Big)\frac{1}{||p_{i^{\prime}}-p_{i}||^{3}}(p_{i} - p_{i^{\prime}})\\
	a_{ii^{\prime}} &=
	\begin{cases}
	\  1, &  if  \  ||p_{i^{\prime}} - p_{i}|| < z_{min} \\
    	 \ 0, &  otherwise. \\
    	\end{cases}
\end{aligned}
\end{equation}
From (\ref{4.initialattractive}), (\ref{4.virtualattractive}), (\ref{4.repulsive}), and (\ref{4.altitude_repulsive}), we have the general control law for the potential field control component:
\begin{equation}\label{4.lowercomponent}
\begin{aligned}
	\ u_{i} &= -\sum_{j \in N_{i}}\nu a_{ij}\Big(\frac{1}{||p_{j} - p_{i}||} - \frac{1}{d}\Big)\frac{1}{||p_{j}-p_{i}||^{3}}(p_{i} - p_{j}) \\
	&- \nu^{\prime} a_{ii^{\prime}}\Big(\frac{1}{||p_{i^{\prime}} - p_{i}||} - \frac{1}{z_{min}}\Big)\frac{1}{||p_{i^{\prime}}-p_{i}||^{3}}(p_{i} - p_{i^{\prime}})\\
	&- (1 - \zeta_{i})k_{r}(p_{i} - p_{r}) -  \zeta_{i} k_{d}(p_{i} - p_{d_{i}}), \\
	\zeta_{i} &= 
	\begin{cases}
    	\ 1, &if  \    Q(t)\cap (B_{i} \cup l_{k,i})\neq \varnothing \\
    	 \ 0, &if  \   otherwise. \\
    	\end{cases}
\end{aligned}
\end{equation}
\noindent
Note that, during the time the UAVs travel to the wildfire region, the coverage control component would not work because the sets $Q(t) \cap B_{i}$ and $Q(t) \cap l_{k,i}$ are initially empty, so $\zeta_{i} = 0$. Upon reaching the waypoint region where the UAVs can sense the fire, $\zeta_{i} = 1$, that would cancel the potential forces that draw the UAVs to the rendezvous point and let the UAVs track the fire fronts growing. The final position of the UAV $i$ will be updated as follows:
\begin{equation}\label{4.finalposition}
\begin{aligned}
	\ p_{i}(t+\Delta t) &= p_{i}(t) + u_{i}\Delta t.\\
\end{aligned}
\end{equation}

\subsection{Stability analysis}

In this section, we study the stability of the proposed control framework. The proof of the stability of the coverage and tracking controller (\ref{4.gradientcomponent}) is similar to the proof in~\cite{schwager2011eyes}. Choose a Lyapunov candidate function $V = H(p_{1}, p_{2}, ..., p_{n})$, where $H$ is the objective function in (\ref{3_coverobjective}). Since $H$ is the area under the FOVs of all UAVs multiplying with point-wise, positive importance index, $H$ is positive definite for all $(p_{1}, p_{2}, ..., p_{n})$. We have:
\begin{equation}\label{4.schwagerstability}
\begin{aligned}
	\ \dot{V} &= [\frac{\partial H}{\partial p_{1}}, \frac{\partial H}{\partial p_{2}}, ..., \frac{\partial H}{\partial p_{n}}]^{T}[\dot{p_{1}}, \dot{p_{2}}, ..., \dot{p_{n}}] \\ 
	&=\sum_{i = 1}^{n} \frac{\partial H}{\partial p_{i}}\dot{p_{i}} =  \sum_{i = 1}^{n} \frac{\partial H}{\partial p_{i}}(-k \frac{\partial H}{\partial p_{i}}) = -k\sum_{i = 1}^{n}(\frac{\partial H}{\partial p_{i}})^{2} \leq 0.\\
\end{aligned}
\end{equation}
Note that $\dot{V} = 0$ if and only if $p_{i} = p_{i}^{*}$ at local minima of $H$ as in (\ref{3_coverobjective}). Therefore, the equilibrium point $p_{i} = p_{i}^{*}$ is asymptotically stable according to Lyapunov stability theorem. The potential field controller is a combination of repulsive and attractive artificial forces in two separatable phases. In the first phase, $\zeta_{i} = 0$, let $p = p_{i} - p_{j}$, $p^{\prime} = p_{i} - p_{i^{\prime}}$, $p_{1} = p_{i} - p_{r}$, and choose a Lyapunov candidate function $V_{1} = \frac{1}{2}p^{2} + \frac{1}{2}{p\prime}^{2} + \frac{1}{2}{p_{1}}^{2}$ which is positive definite, radially unbounded. We have:
\begin{equation}\label{4.potentialstability}
\begin{aligned}
	\ \dot{V_{1}} &= p\dot{p} + p\prime \dot{p\prime} + p_{1}\dot{p_{1}} = p u_{i}^{rep1} + p\prime u_{i}^{rep2}+ p_{1} u_{i}^{r} \\ 
	&=   p_{1}(-kp_{1}) - \sum_{j \in N_{i}}\nu a_{ij}\Big(\frac{1}{||p||} - \frac{1}{d}\Big)\frac{1}{||p||^{3}}p^{2}\\ 
	&-  \nu^{\prime} a_{ii^{\prime}}\Big(\frac{1}{||p\prime||} - \frac{1}{z_{min}}\Big)\frac{1}{||p\prime||^{3}}p\prime^{2}  \leq 0,\\
\end{aligned}
\end{equation}
since $\frac{1}{||p||} - \frac{1}{d} > 0$ and $\frac{1}{||p\prime||} - \frac{1}{z_{min}} > 0$. $V_{1} = 0$ if and only if at equilibrium points. Therefore, the equilibrium points $p_{i} = p_{r}, p_{i} = p_{j}, p_{i} = p_{i^{\prime}}$ are global asymptotically stable. The proof for second phase, $\zeta_{i} = 1$, is similar. In conclusion, the two controllers are asymptotically stable.

\begin{algorithm}
\DontPrintSemicolon % Some LaTeX compilers require you to use \dontprintsemicolon instead
\KwIn{Real-time localization the UAV $p_{i}$ and other neighbor UAVs $p_{j}, j \in N$. Heat intensity of each point $I(q)$ under the FOV}
\KwOut{New position $p_{i}$}

\For{$i =  1:N$}{
	Locate FOV of UAV $i$ and discretize them to get the set of points $\hat{B_{i}}$ and its four edges $\hat{l_{k,i}}, k = 1:4$

	Check if this point is on the fire $Q(t)$ to compute $Q(t) \cap \hat{B_{i}}$ and  $Q(t) \cap\hat{l_{k,i}}$

	\uIf{$Q(t)\cap B_{i} = \varnothing$}{
		Calculate $u_{i}^{rep1}$, $u_{i}^{rep2}$ according to (\ref{4.repulsive}) and (\ref{4.altitude_repulsive})\\
		Calculate $u_{i}$ according to (\ref{4.lowercomponent}):
		\begin{equation*}
		\begin{aligned}
		\ u_{i} =  u_{i}^{rep1} + u_{i}^{rep2} - k_{r}(p_{i} - p_{r})\\
		\end{aligned}
		\end{equation*}
		Update:
		$p_{i}(t+\Delta t) = p_{i}(t) -u_{i}\Delta t$\\
	}
	\Else{
		
		\For{$q \in Q(t) \cap \hat{B_{i}} \& Q(t) \cap\hat{l_{k,i}}$}{
		Compute $f_{N_{q}}$ and $ f_{N_{q\setminus i}}$\\
		Estimate $\Delta I(q) = I_{max} - I(q)$\\
		}

		Compute:\\
		\begin{equation*}
		\begin{aligned}
		\frac{\Delta H}{\Delta  c_{i}} = \sum_{k=1}^{4}\sum_{q \in \hat{Q(t) \cap l_{k,i}}}(f_{N_{q}} - f_{N_{q}\setminus i})n_{k}\kappa\Delta I(q)\Delta q \\
		\frac{\Delta H}{\Delta  z_{i}} = \sum_{k=1}^{4}\sum_{q \in \hat{Q(t) \cap l_{k,i}}}(f_{N_{q}} - f_{N_{q}\setminus i})tan \theta^{T}n_{k}\\
		\ \kappa\Delta I(q)\Delta q\\ 
		- \sum_{q \in \hat{Q(t) \cap B_{i}}} \frac{2f_{N_{q}}^{2}}{ \frac{S_{1}}{b^{2}}(b-z_{i})^{3}}\kappa\Delta I(q)\Delta q \\
		\end{aligned}
		\end{equation*}

		Update:
		$p_{d_{i}}(t+\Delta t) = p_{d_{i}}(t) -  (k_{c}\frac{\Delta H}{\Delta c_{i}}, k_{z}\frac{\Delta H}{\Delta z_{i}})\Delta t$
		Calculate $u_{i}^{rep1}$, $u_{i}^{rep2}$ according to (\ref{4.repulsive}) and (\ref{4.altitude_repulsive})\\
		Go to desired position $p_{d_{i}}$ according to (\ref{4.lowercomponent}):
		\begin{equation*}
		\begin{aligned}
		\ u_{i} =  u_{i}^{rep1} + u_{i}^{rep2} - k_{d}(p_{i} - p_{d_{i}})\\
		\end{aligned}
		\end{equation*}
		Update:
		$p_{i}(t+\Delta t) = p_{i}(t) -u_{i}\Delta t$\\
	}
}

\caption{{\sc Distributed wildfire tracking control.}}
\label{algo1}
\end{algorithm}

\begin{figure*}[h]
 \centering
	\subfloat[t = 1000]{\includegraphics[width=0.25\textwidth]{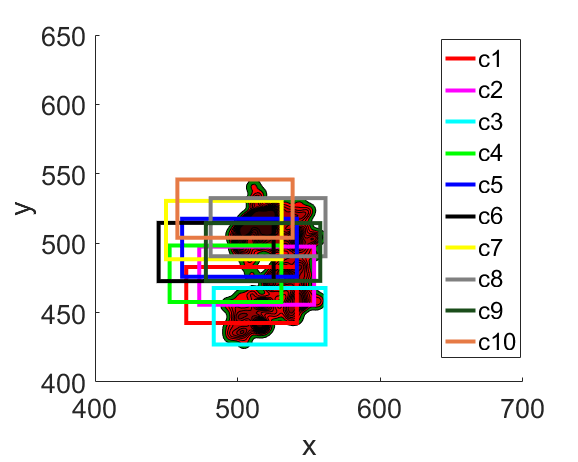}}
	\subfloat[t = 3000]{\includegraphics[width=0.25\textwidth]{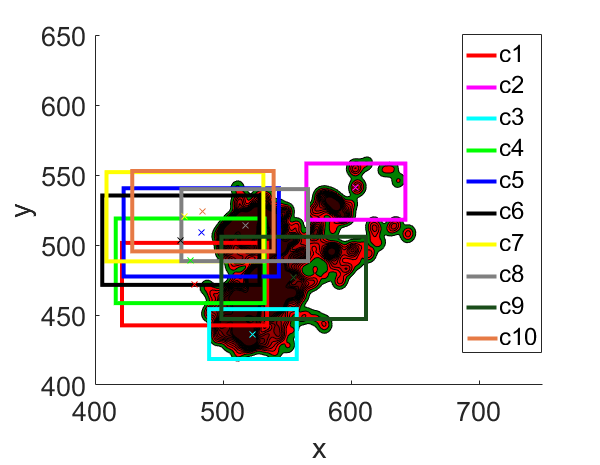}}
	\subfloat[t=  4000]{\includegraphics[width=0.25\textwidth]{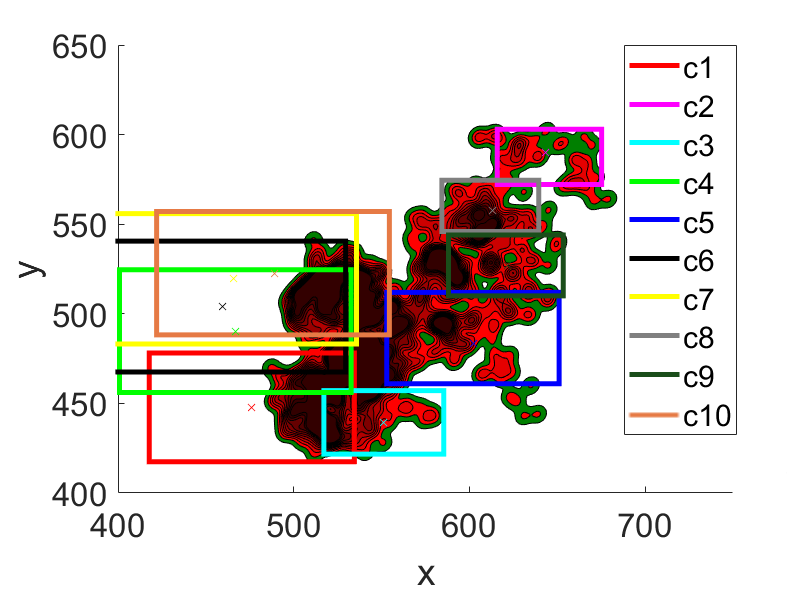}}
	\subfloat[t = 6000]{\includegraphics[width=0.25\textwidth]{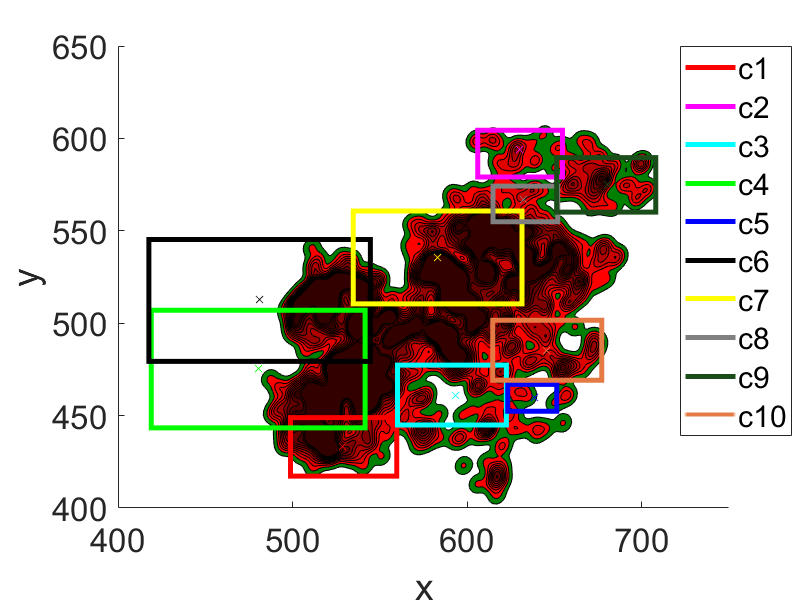}}
  \caption{Simulation result shows the FOV of each UAV on the ground in a) t = 1000, b) t = 3000, c) t= 4000, and d) t = 6000. The UAVs obviously followed the newly developed fire front propagation.}
  \label{F.Simulationc}
\vspace{-0pt}
\end{figure*}

\begin{figure*}[h]
 \centering
	\subfloat[t = 1000]{\includegraphics[width=0.25\textwidth]{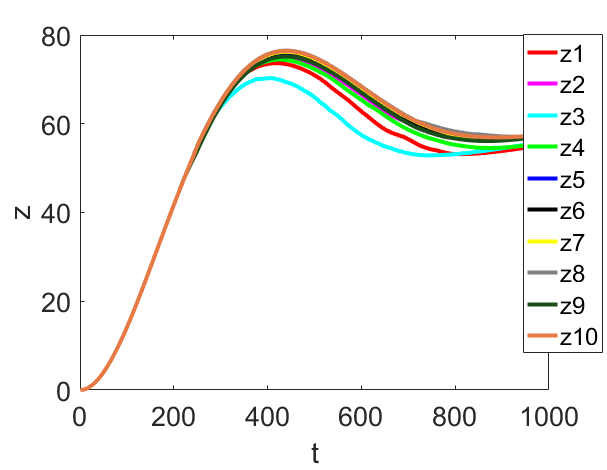}}
	\subfloat[t = 3000]{\includegraphics[width=0.25\textwidth]{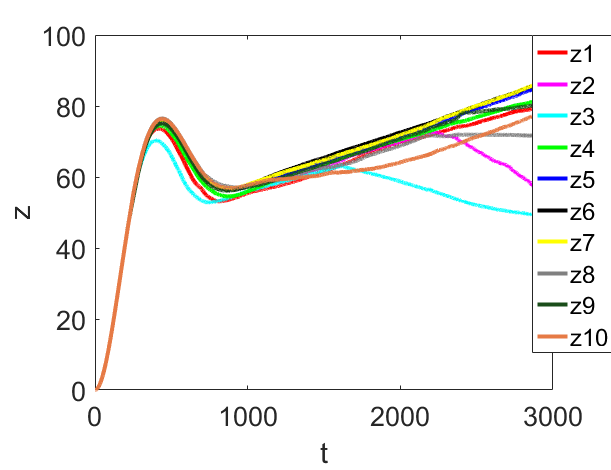}}
	\subfloat[t=  4000]{\includegraphics[width=0.25\textwidth]{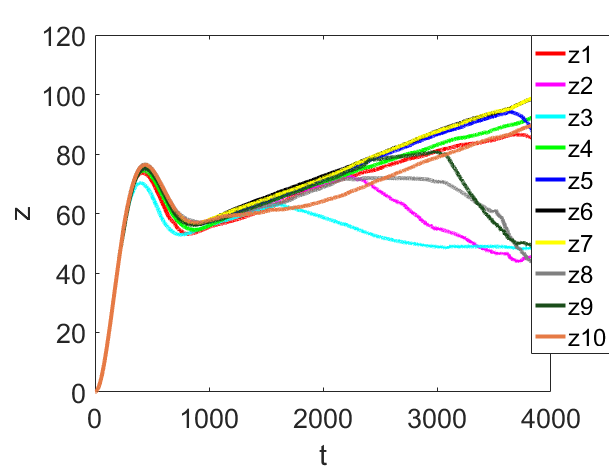}}
	\subfloat[t = 6000]{\includegraphics[width=0.25\textwidth]{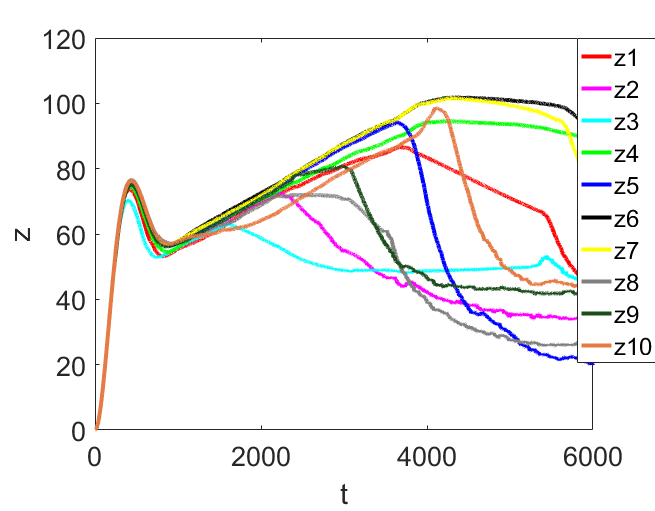}}
  \caption{Plots showing the altitude of each UAV from the ground in a) t = 1000, b) t = 3000, c) t= 4000, and d) t = 6000. The UAVs change altitude from $z_{i} \approx 60$ to different altitudes, making the area of the FOV of each UAV is different.}
  \label{F.Simulationz}
\vspace{-0pt}
\end{figure*}

\begin{figure*}[htb!]
 \centering
  \includegraphics[width=0.9\textwidth]{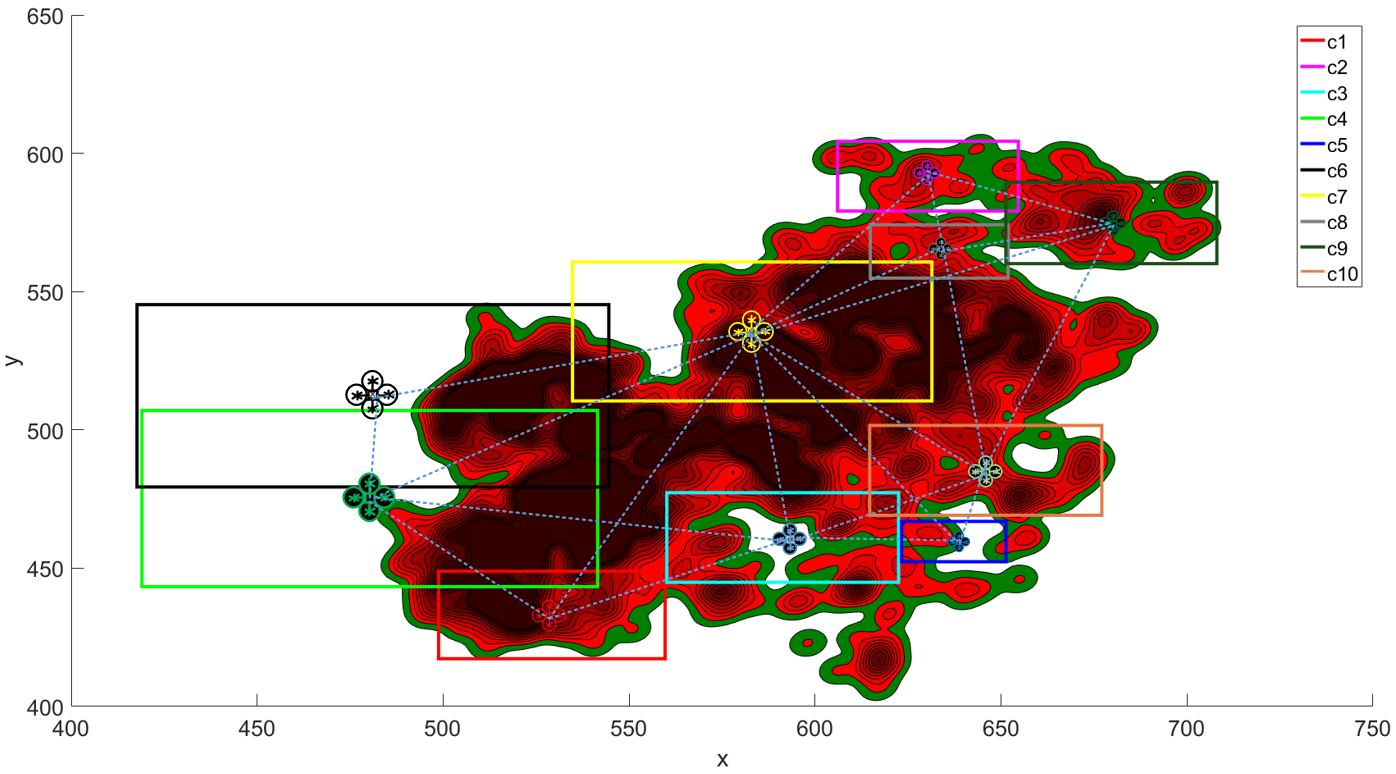}
  \caption{Rendering showing UAV positions and FOV during wildfire tracking, showing that the UAVs attempted to follow the fire front propagation, with greater focus on newly developed fire front.}
  \label{F.UAVFOV}
\vspace{0 cm}
\end{figure*}

\begin{figure*}[htb!]
 \centering
  \includegraphics[width=1.1\textwidth]{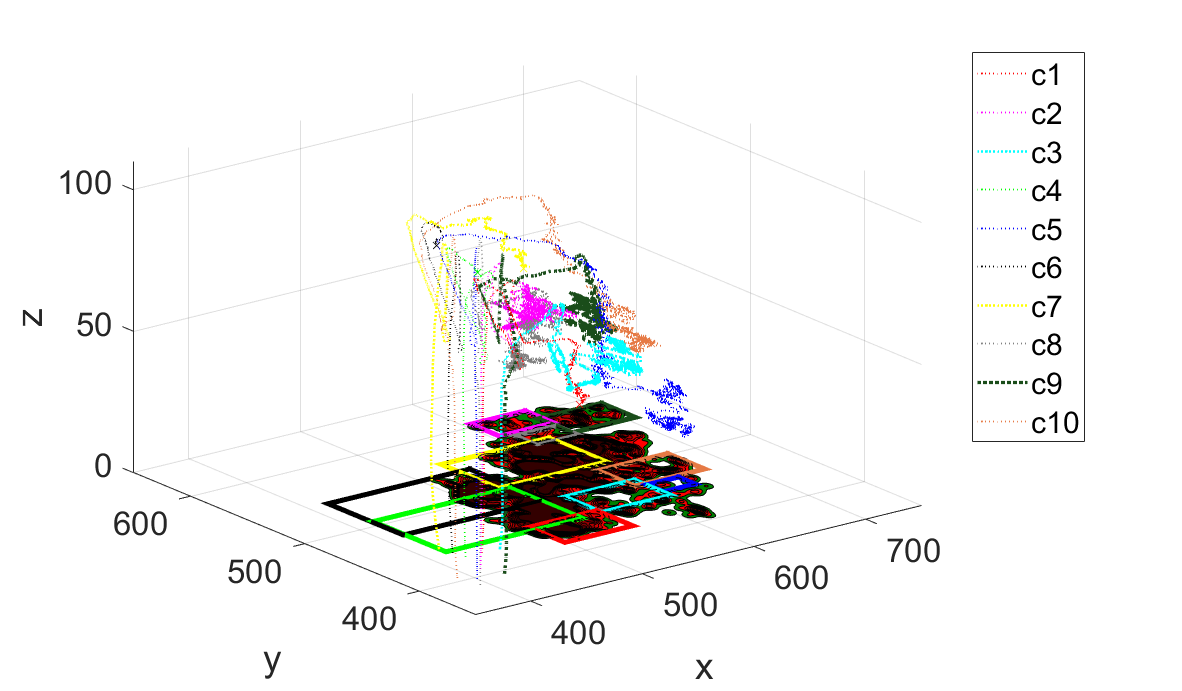}
  \caption{3D representation of the UAVs showing the trajectory of each UAV in 3-dimensions while tracking the wildfire spreading north-east, and their current FOV on the ground.}
  \label{F.Simulation3D}
\vspace{-0pt}
\end{figure*}

\subsection{Overall algorithm for decentralized control of UAVs}

We implemented the control strategy for the UAVs in a distributed manner as summarized in Algorithm \ref{algo1}. Each UAV needs to know its position from localization using means such as GPS+IMU fusion with Extended Kalman Filter (EKF) at each time step~\cite{la2013mechatronic, la2014autonomous, la2017development}. They can also communicate with other UAVs within the communication range to get their positions. Each UAV must also be able to read the heat intensity of any point under its FOV from the sensor. The coverage and tracking control component will calculate the new position for the UAV in each loop. To move to a new position, a UAV will use the potential field control component which takes the new position as their input. To calculate the integrals in (\ref{4.gradientcomponent}), we need to discretize the rectangular FOV of a UAV and its four edges in to a set of points, with $\Delta q$ is either the length of a small line in each edge or the area of a small square. The integrals can then be transformed into the sum of all the small particles.

When activated, the UAV will first discretize its rectangular FOV into sets of points of a grid (line 3). These points will be classified into sets of edges $\hat{l_{k}}, k = 1:4$, and a set for the area inside the FOV $\hat{B_{i}}$, together with the value of $\Delta q$ associated with each set. Then the UAV would read the intensity level of each point, $I(q)$, of these sets to determine if the point is currently in the fire or not, and form the set $Q(t) \cap \hat{B_{i}}$ and $Q(t) \cap\hat{l_{k,i}}$. If the sets $Q(t) \cap \hat{B_{i}}$ and $Q(t) \cap\hat{l_{k,i}}$ are not empty, then it would go to the rendezvous point (line 6). This will help the UAV to go to the right place in the initialization phase, as well as help the UAVs not to venture completely out of the fire. If at least one set is not empty, it will then identify the set $N_{q}$ and $N_{q\setminus i}$ by testing with equation (\ref{3_coveragecondition}), and compute $\Delta I(q)$, $f_{N_{q}}$ and $ f_{N_{q\setminus i}}$ as in (\ref{3_Schwagersobjective}), for every point in $Q(t) \cap \hat{B_{i}}$ and $Q(t) \cap\hat{l_{k,i}}$). The integrals in (\ref{4.gradientcomponent}) then can be calculated, and the new position $p_{d_{i}}$ is then updated as in (\ref{4.virtualposition}).

\section{Simulation}
Our simulation was conducted in a Matlab environment. We started with 10 UAVs on the ground ($z_{i} = 0$) from a fire fighting center with initial location arbitrarily generated around $[300, 300]^{T}$. The safe distance was $d = 10$, and the safe altitude was $z_{min} = 15$. The UAVs were equipped with identical cameras with focal length $b = 10$, area of one pixel $S_{1} = 10^{-4}$, half-angles $\theta_{1}=30^{\circ}$, $\theta_{2}=45^{\circ}$. We chose parameter $m = 1.5 \time 10^{-5}$ to avoid zero division as in (\ref{3_Schwagersobjective}). The intensity sensitivity range of each camera was $[0.005, 0.1]^{T}$, and $\kappa = 1$. The wildfire started with five initial fire front points near $[500, 500]^{T}$. The regulated mid-flame wind speed magnitude followed a Gaussian distribution with $\mu = 5 mph$ and $\sigma = 2$. The wind direction azimuth angle $\Theta$ also followed a Gaussian distribution with $\mu = \frac{\pi}{8}$ and $\sigma = 1$. The UAVs had a communication range $r = 500$.

We conducted tests in two different scenarios. In the first test, the UAVs performed wildfire coverage with specific focus on border of the fire, while in the second one, the UAVs have no specific focus on the border. In both of the two scenarios, the coverage and tracking controller parameters were $k_{c} = 10^{-9}$, $k_{z} = 2 \time 10^{-10}$, while the potential field controller parameters were $k_{r} = k_{d} = 0.06$, $\nu = 2.1$ and $\nu \prime = 10^{3}$. The simulation parameters, presented in Table \ref{tab:sce1} and Table \ref{tab:sce2}, were selected after some experiments.

\subsection{Scenario: Wildfire coverage with specific focus on border of the fire}

\begin{table}
\begin{center}
  \caption{Simulation parameters for wildfire coverage with specific focus on border of the fire}
  \begin{tabular}{{|m{1.7cm}|m{1.7cm}|m{1.7cm}|m{1.7cm}|}}
\hline
\multicolumn{2}{|c|}{Wind direction angle $\Theta$} & \multicolumn{2}{|c|}{Wind speed magnitude $U$}\\
\hline
 $\mu = \frac{\pi}{8} rad$ &  $\sigma = 1$ &  $\mu = 5 mph$ &  $\sigma = 2$\\
\hline
\multicolumn{4}{|c|}{Camera and sensing parameters}\\
\hline
$b = 10$ & $S_{1} = 10^{-4}$ & $\theta_{1}=30^{\circ}$ & $\theta_{2}=45^{\circ}$\\
\hline
$m=1.5 \time 10^{-5}$ & $I_{min}=0.005$ & $I_{max}=0.1$ & $\kappa=1$\\
\hline
\multicolumn{2}{|c|}{Coverage \& tracking}  & \multicolumn{2}{|c|}{Safe distance}\\
\hline
$k_{c} = 10^{-9}$ & $k_{z} = 2 \time 10^{-10}$ &  $d = 10$ &  $z_{min} = 15$\\
\hline
\multicolumn{4}{|c|}{Potential field controller's parameters}\\
\hline
$k_{r} = 0.06$ & $k_{d} = 0.06$ & $\nu = 2.1$ & $\nu \prime = 10^{3}$\\
\hline
\end{tabular}
\label{tab:sce1}
\end{center}
\vspace{-10pt}
\end{table}

\begin{figure*}[h]
 \centering
	\subfloat[t = 1000]{\includegraphics[width=0.25\textwidth]{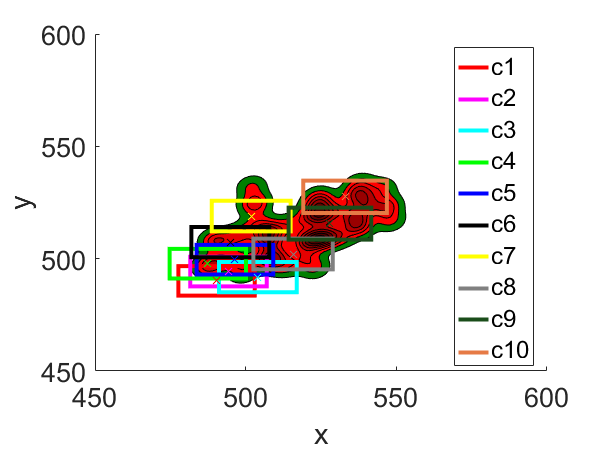}}
	\subfloat[t = 3000]{\includegraphics[width=0.25\textwidth]{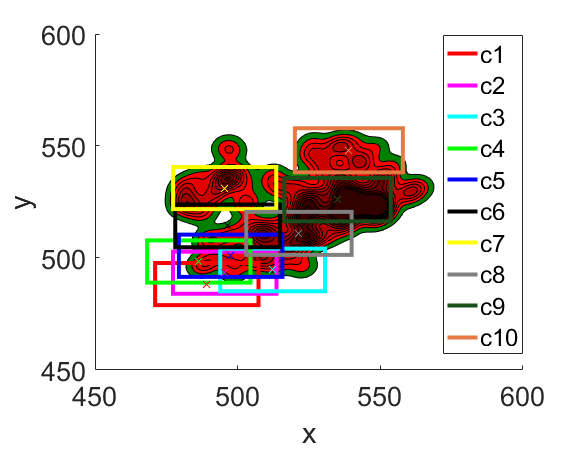}}
	\subfloat[t=  4000]{\includegraphics[width=0.25\textwidth]{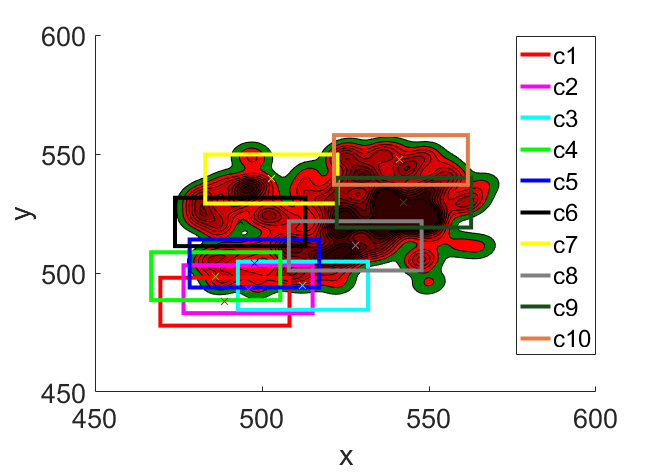}}
	\subfloat[t = 6000]{\includegraphics[width=0.25\textwidth]{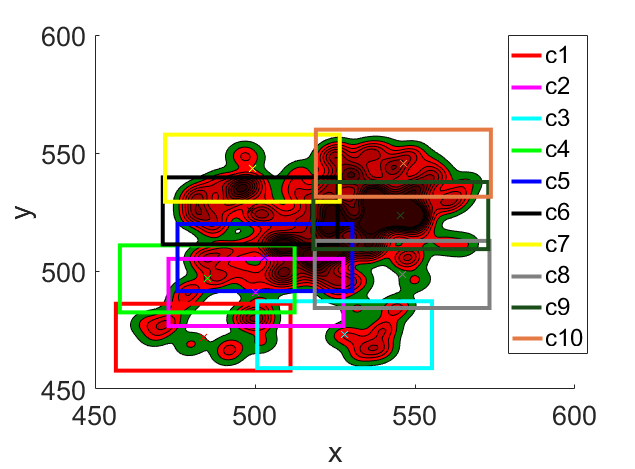}}
  \caption{Simulation result shows the FOV of each UAV on the ground in a) t = 1000, b) t = 3000, c) t= 4000, and d) t = 6000. The whole wildfire got covered, with no specific focus.}
  \label{F.PhiconstSimulationc}
\vspace{-0pt}
\end{figure*}
%Figure 11: Phi Const z simulation
\begin{figure*}[h]
 \centering
	\subfloat[t = 1000]{\includegraphics[width=0.25\textwidth]{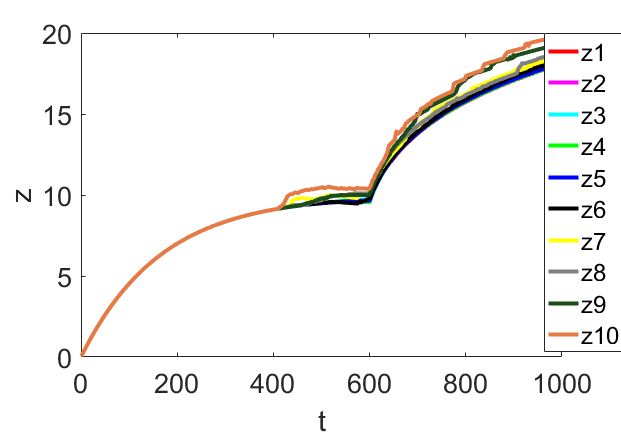}}
	\subfloat[t = 3000]{\includegraphics[width=0.25\textwidth]{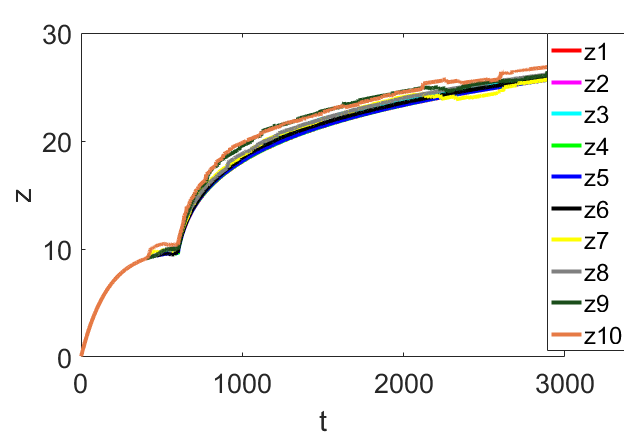}}
	\subfloat[t=  4000]{\includegraphics[width=0.25\textwidth]{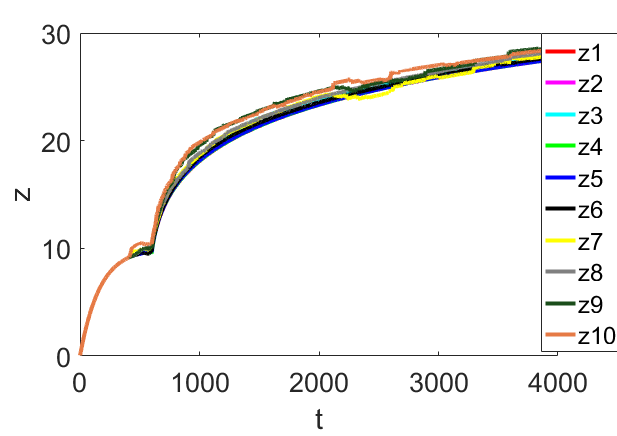}}
	\subfloat[t = 6000]{\includegraphics[width=0.25\textwidth]{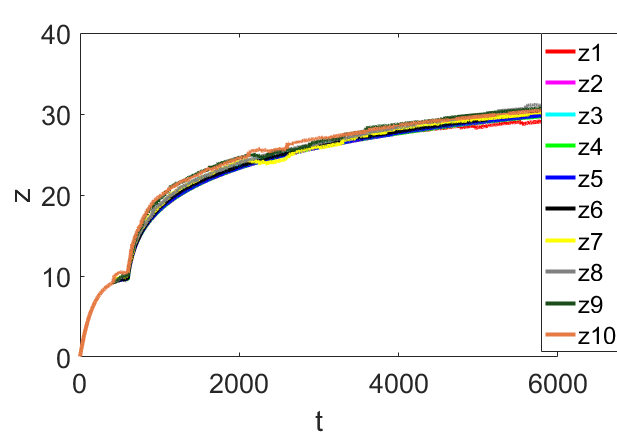}}
  \caption{Plot shows the altitude of each UAV on the ground in a) t = 1000, b) t = 3000, c) t= 4000, and d) t = 6000. Since they were not focusing on the border of the fire, the altitudes of the UAVs were almost equal.}
  \label{F.PhiconstSimulationz}
\vspace{-0pt}
\end{figure*}

The main parameters for the simulation were given in Table \ref{tab:sce1}. We ran simulations in MATLAB for 6000 time steps which yielded the result as shown in Figures \ref{F.Simulationc} and \ref{F.Simulationz}. The UAVs came from the ground at $t = 0$ (Figure \ref{F.Simulationz}), and drove toward the wildfire region. The initial rendezvous point was $p_{r} = [500, 500, 60]^{T}$. Upon reaching the region near the initial rendezvous point at $[500, 500]^{T}$, the UAVs spread out to cover the entire wildfire (Figure \ref{F.Simulationc}-a). As the wildfire expanded, the UAVs fragment and follow the fire border regions (Figure \ref{F.Simulationc}-b, c, d). Note that the UAVs may not cover some regions with intensity $I = I_{max}$ (represented by black-shade color). Some UAVs may have low altitude if they cover region with small intensity $I$ (for example, UAV 5 in this simulation). The UAVs change altitude from $z_{i} \approx 60$ (Figure \ref{F.Simulationz}-a) to different altitudes (Figure \ref{F.Simulationz}-b, c, d), hence the area of the FOV of each UAV is different. It is obvious to notice that the UAVs attempted to follow the fire front propagation, hence satisfying the tracking objective. Figure \ref{F.UAVFOV} indicates the position of each UAV and its respective FOV in the last stage $t =  6000$. UAVs that are physical neighbors are connected with a dashed blue line. We can see that most UAVs have sensing neighbors. Figure \ref{F.Simulation3D} shows the trajectory of each UAV in 3-dimensions while tracking the wildfire spreading north-east, and their current FOV on the ground.

\subsection{Scenario: Normal wildfire coverage}

\begin{table}
\begin{center}
  \caption{Simulation parameters for normal wildfire coverage with no specific focus}
  \begin{tabular}{{|m{1.7cm}|m{1.7cm}|m{1.7cm}|m{1.7cm}|}}
\hline
\multicolumn{2}{|c|}{Wind direction angle $\Theta$} & \multicolumn{2}{|c|}{Wind speed magnitude $U$}\\
\hline
 $\mu = \frac{\pi}{8} rad$ &  $\sigma = 1$ &  $\mu = 5 mph$ &  $\sigma = 2$\\
\hline
\multicolumn{4}{|c|}{Camera and sensing parameters}\\
\hline
$b = 10$ & $S_{1} = 10^{-4}$ & $\theta_{1}=30^{\circ}$ & $\theta_{2}=45^{\circ}$\\
\hline
$m = 1.5 \time 10^{-5}$ & $I_{min} = N/A$ & $I_{max} = N/A$ & $\kappa = 10^{-3}$\\
\hline
\multicolumn{2}{|c|}{Coverage \& tracking}  & \multicolumn{2}{|c|}{Safe distance}\\
\hline
$k_{c} = 10^{-9}$ & $k_{z} = 2 \time 10^{-10}$ &  $d = 10$ &  $z_{min} = 15$\\
\hline
\multicolumn{4}{|c|}{Potential field controller's parameters}\\
\hline
$k_{r} = 0.06$ & $k_{d} = 0.06$ & $\nu = 2.1$ & $\nu \prime = 10^{3}$\\
\hline
\end{tabular}
\label{tab:sce2}
\end{center}
\vspace{-10pt}
\end{table}

In this simulation scenario, we demonstrate the ability of the group of UAVs to cover the spreading fire with no specific focus. The main simulation parameters for this scenario were given in Table \ref{tab:sce2}. The control strategy and parameters were the same as in the previous scenario, except there was no special interest in providing higher-resolution images of the fire border, therefore, equation \ref{3_phi} became $\phi(q) = \kappa$. The initial rendezvous point was $p_{r} = [500, 500, 10]^{T}$. As we can see in Figures \ref{F.PhiconstSimulationc} and \ref{F.PhiconstSimulationz}, the UAVs covered the fire spreading very well, with no space uncovered. Since they were not focusing on the border of the fire, the altitudes of the UAVs were almost equal.

\section{Conclusion}

In this paper, we presented a distributed control design for a team of UAVs that can collaboratively track a dynamic environment in the case of wildfire spreading. The UAVs can follow the border region of the wildfire as it keeps expanding, while still trying to maintain coverage of the whole wildfire. The UAVs are also capable of avoiding collision, maintaining safe distance to fire level, and flexible in deployment. The application could certainly go beyond the scope of wildfire tracking, as the system can work with any dynamic environment, for instance, oil spilling or water flooding. In the future, more work should be considered to research about the hardware implementation of the proposed controller. For example, we should pay attention to the communication between the UAVs under the condition of constantly changing topology of the networks, or the sensing endurance problem in hazardous environment. Also, we would like to investigate the relation between the speed of the UAVs and the spreading rate of the wildfire, and attempt to synchronize it. Multi-drone cooperative sensing~\cite{nguyen2017collaborative, La_SMCB_2013, La_SMCA_2015}, cooperative control~\cite{nguyen2017formation, La_ICRA2010, La_SMC2009}, cooperative learning~\cite{La_TCST_2015, La_CYBER2013}, and user interface design~\cite{ahmedsiddiqui2017development} for wildland fire mapping will be also considered.

\section*{ACKNOWLEDGMENT}
This material is based upon work supported by the National Aeronautics and Space Administration (NASA) Grant No. NNX15AI02H issued through the Nevada NASA Research Infrastructure Development Seed Grant, and the National Science Foundation \#IIS-1528137.

\bibliographystyle{IEEEtran}
\bibliography{SMCA2018_bibliography}
\vspace{-30pt}
\begin{IEEEbiography}[{\includegraphics[width=1in,keepaspectratio]{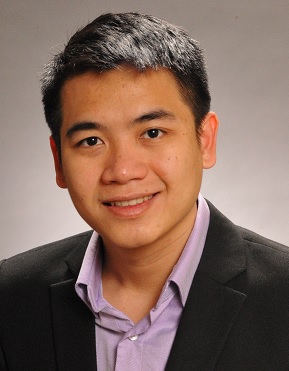}}]{Huy X. Pham} received his B.S. degree in Electrical Engineering from Hanoi University of Science and Technology, Hanoi, Vietnam in 2011. He received M.S. degree in Industrial and Systems Engineering from the University of Oklahoma, Norman, OK, USA, in 2015. From 2017, Mr. Huy worked in the Advanced Robotics and Automation (ARA) Lab, University of Nevada, Reno, to conduct research for pursuing his PhD degree. His research interests include robotics and control systems, multi-robot systems, and collaborative learning. \
\vspace{-30pt}
\end{IEEEbiography}
\begin{IEEEbiography}[{\includegraphics[width=1in,height=1.25in,clip,keepaspectratio]{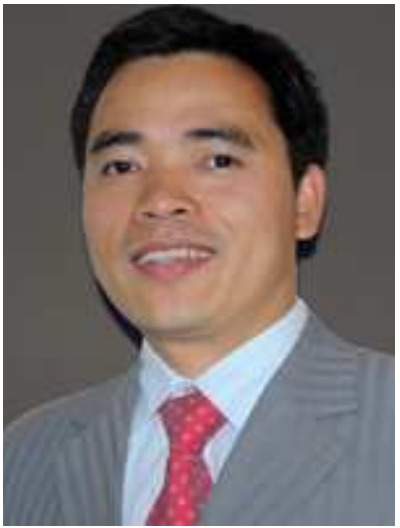}}]{Hung M. La} 
(IEEE SM'2014, M'2009) received his B.S. and M.S. degrees in Electrical Engineering from Thai Nguyen University of Technology, Thai Nguyen, Vietnam, in 2001 and 2003, respectively, and his Ph.D. degree in Electrical and Computer Engineering from Oklahoma State University, Stillwater, OK, USA, in 2011.
He is the Director of the Advanced Robotics and Automation (ARA) Lab, and Assistant Professor of the Department of Computer Science and Engineering, University of Nevada, Reno, NV, USA.  From 2011 to 2014, he was a Post Doctoral research fellow and then a Research Faculty Member at the Center for Advanced Infrastructure and Transportation, Rutgers University, Piscataway, NJ, USA.  Dr. La is an Associate Editor of the IEEE Transactions on Human-Machine Systems,  and Guest Editor of International Journal of Robust and Nonlinear Control. \
\vspace{-30pt}
\end{IEEEbiography}
\begin{IEEEbiography}[{\includegraphics[width=1in,height=1.25in,clip,keepaspectratio]{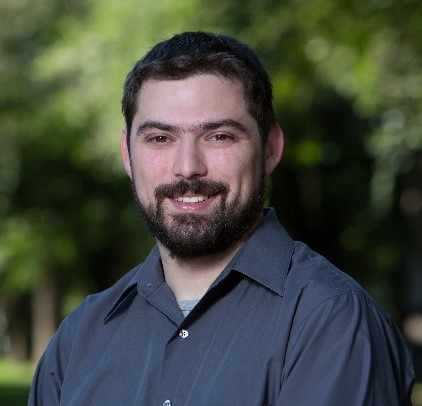}}]{David Feil-Seifer} (IEEE M'2006) received a B.S. degree in Computer Science from the University of Rochester, Rochester, NY in 2003. He received M.S. and Ph.D. degrees in Computer Science from  the University of Southern California, Los Angeles, CA in 2007 and 2012, respectively. Dr. Feil-Seifer is an Assistant Professor and Director of the Socially Assistive Robotics Group (SARG) in the Department of Computer Science \& Engineering at the University of Nevada, Reno, NV, USA since July, 2013. From 2011-2013, he was a Postdoctoral Associate in the Yale University Computer Science Department, New Haven, CT, USA. He has authored over 45 papers published in major journals, book chapters, and international conference proceedings. His current research interests include Human-Robot Interaction and Socially Assistive Robotics. Dr. Feil-Seifer is Managing Editor of the ACM Transactions on Human-Robot Interaction. \
\vspace{-50pt}
\end{IEEEbiography}
\begin{IEEEbiography}[{\includegraphics[width=1in,height=1.25in,clip,keepaspectratio]{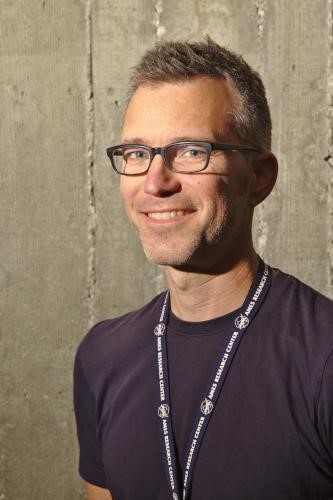}}]{Matthew C. Deans} received B.S. and M.S. degrees in Electrical Engineering and a B.S. in Engineering Physics from Lehigh University in 1994, 1995, and 1996; and a Ph.D. in Robotics from Carnegie Mellon University in 2002. Dr. Deans is the Deputy Director of the Intelligent Robotics Group at NASA Ames Research Center in Silicon Valley, CA, USA.  His research interests include computer vision, field robotics, robot operations, and robotic planetary space flight missions. \
\vspace{-30pt}
\end{IEEEbiography}

\end{document}